\documentclass[lettersize,journal]{IEEEtran}

\usepackage[utf8]{inputenc}
\usepackage[T1]{fontenc}
\usepackage{microtype,newtxtext,newtxmath} 
\usepackage{mathtools,gensymb,url} 
\usepackage{subcaption}%
\usepackage[font=footnotesize,textfont=footnotesize]{subcaption}
\usepackage{longtable,booktabs,array}
\usepackage[backend=biber, sorting=none, style=ieee, maxnames=5]{biblatex}
\usepackage{acronym}
\usepackage{graphicx}
\usepackage{hyperref}
\usepackage{siunitx}
\hypersetup{colorlinks=true, breaklinks=true, 
  urlcolor=blue, linkcolor=blue,anchorcolor=blue,citecolor=blue,
  hypertexnames=true, final=true, 
  pdfpagemode = UseNone, 
  pdfauthor = {},
  pdftitle = {},   
  pdfsubject = {},
  pdfkeywords = {}
}

\graphicspath{{img/} {figs/} {./}}
\DeclareGraphicsExtensions{.pdf,.png,.jpg,.mps}

\IEEEdisplaynontitleabstractindextext

\DeclareSourcemap{
  \maps[datatype=bibtex]{
    \map[overwrite=true]{
      \step[fieldset=url, null]
      \step[fieldset=doi, null]
      \step[fieldset=eprint, null]
       \step[fieldset=month, null]  
    }
  }
}

\usepackage{threeparttable}
\usepackage{pifont}
\newcommand{\cmark}{\ding{51}}
\newcommand{\xmark}{\ding{55}}
\usepackage{fancyhdr}
\fancypagestyle{headnote}{%
   \fancyhf{} 
   
   \fancyhead[C]{This work has been submitted to the IEEE for possible publication. Copyright may be transferred without notice, after which this version may no longer be accessible.}
}%
\makeatletter
\let\ps@IEEEtitlepagestyle\ps@headnote

\makeatother

\begin{document}
\title{EchoSpike Predictive Plasticity: An Online Local Learning Rule for Spiking Neural Networks}
\author{
Lars Graf\textsuperscript{1}, Zhe Su$^{\ast}$\textsuperscript{,2}, Giacomo Indiveri\textsuperscript{2}

\textsuperscript{1}\textit{ETH Zurich} Zurich, Switzerland\\
\textsuperscript{2}\textit{Institute of Neuroinformatics} \textit{University of Zurich and ETH Zurich} Zurich, Switzerland \\

\thanks{*Corresponding author Zhe Su, zhesu@ini.uzh.ch}
}

\maketitle
\begin{abstract}
The drive to develop artificial neural networks that efficiently utilize resources has generated significant interest in bio-inspired Spiking Neural Networks (SNNs). These networks are particularly attractive due to their potential in applications requiring low power and memory. This potential is further enhanced by the ability to perform online local learning, enabling them to adapt to dynamic environments. This requires the model to be adaptive in a self-supervised manner. While self-supervised learning has seen great success in many deep learning domains, its application for online local learning in multi-layer SNNs remains underexplored. In this paper, we introduce the ``EchoSpike Predictive Plasticity'' (ESPP) learning rule, a pioneering online local learning rule designed to leverage hierarchical temporal dynamics in SNNs through predictive and contrastive coding. We validate the effectiveness of this approach using benchmark datasets, demonstrating that it performs on par with current state-of-the-art supervised learning rules. The temporal and spatial locality of ESPP makes it particularly well-suited for low-cost neuromorphic processors, representing a significant advancement in developing biologically plausible self-supervised learning models for neuromorphic computing at the edge.
\end{abstract}


\section{Introduction}
\label{sec:introduction}
As technology advances, there is an increasing need to process sensory data directly at the point of collection, known as edge computing, rather than relying on distant servers. This transition requires applications that are not only close to the data source but also energy-efficient, fast in processing, robust, and adaptable to changes. The primary challenge is developing processors capable of continuously handling real-time sensor data, extracting necessary information while minimizing energy consumption.

One promising solution involves the use of event-based spiking neural networks (SNNs). Neuromorphic processors employ event-based architectures that integrate in-memory computing with computational concepts inspired by the human brain, particularly for deploying SNNs. These processors are unique due to their asynchronous, spike-based processing, which offers exceptional efficiency and low power consumption. This makes them particularly suitable for managing analog signals over noisy channels, improving noise resistance, and reducing both bandwidth and energy requirements.

On the other hand, exploring online local learning to empower edge computing devices with continual adaptation to their specific environments is an active area of research.
This is particularly important for edge-computing applications, as conventional neural network training algorithms are not well suited for implementation in resource-constrained setups that have limited memory, power, and size requirements.
However, despite their potential advantages, most online local learning rules proposed so far have some critical drawbacks: (1) Cannot resort to large data-sets, due to the scarcity of labeled data typical for edge applications (whereas there exists a wealth of untapped unlabeled data); (2) Require many more training iterations compared to conventional offline training algorithms; (3) Poor scalability for deep neural networks.

To address these limitations, we propose a novel local learning rule algorithm, the ``EchoSpike Predictive Plasticity'' (ESPP) which, inspired by the field of self-supervised learning (SSL), capitalizes on the advantages of spikes in SNNs.
We demonstrate that ESPP effectively addresses the issues of training cost and scalability in local supervised learning rules, validating its performance on two neuromorphic datasets: Spiking Heidelberg Digits (SHD)~\cite{heidelberg} and Neuromorphic-MNIST (N-MNIST)~\cite{nmnist}.
We argue that ESPP is particularly well suited for low-power implementation on neuromorphic hardware, making it a promising candidate for real-world applications at the edge where efficiency and adaptability are paramount.
\section{Related Work}
\label{sec:related-work}
\subsection{Self-supervised learning}
SSL takes advantage of unlabeled data to learn useful representations.
This approach has shown remarkable success in standard Artificial Neural Networks (ANNs).
Its applications in natural language processing (NLP) are particularly noteworthy.
For instance, models like BERT~\cite{BERT} and GPT~\cite{GPT} have revolutionized NLP by using SSL to understand context, semantics, and language structures with minimal reliance on labeled data.

There are intriguing parallels between SSL and learning in biological neural networks (BNNs).
In biological systems, learning often occurs without explicit external supervision, relying instead on internally generated signals and feedback mechanisms.
BNNs are able to efficiently learn a large variety of tasks.
This is especially impressive considering that, in contrast to ANNs and backpropagation, neurons learn by utilizing only information that is locally available.

SNNs, which closely mimic the dynamics of biological neurons, including their temporal spiking behavior, have been shown to be difficult to train, especially in deep neural networks.
Recent efforts have made progress in improving the performance of learning in deep SNNs through backpropagation~\cite{Neftci_etal19, superspike, Lee_etal16}.
However, critical difficulties with training deep SNNs remain, especially when having to use local learning rules.

\subsection{Supervised local learning}
~\cite{eprop} presents a novel time-local learning rule for recurrent spiking neural networks (RSNNs): e-prop, which resolves the temporal credit assignment problem by leveraging so called eligibility traces for efficient online learning. ~\cite{ETLP} introduces an event-based, three-factor local plasticity rule (ETLP), designed for neuromorphic hardware, enabling online learning by integrating eligibility traces from e-prop with direct random target projection (DRTP)~\cite{DRTP}. This leads to a learning rule that is both local in time and space. ~\cite{osttp} proposes a similar method called online spatio-temporal learning with target projection (OSTTP). Instead of building upon e-prop, OSTTP builds upon their previous work (OSTL~\cite{ostl}), resulting in a slightly different eligibility trace. However, both OSTTP and ETLP have the previously mentioned issues.

\subsection{Self-supervised local learning}
Given the success of SSL in both ANNs and BNNs, it is increasingly evident that SSL can play a pivotal role in improving local learning in SNNs.
An interesting approach for utilizing self-supervised local learning is latent predictive learning (LPL)~\cite{combination}, which combines predictive coding with hebbian learning.
In LPL, each layer can be interpreted as making predictions about its own future activity.
In contrast to making predictions directly on raw input data in the input space, predictions in a latent space are based on higher-level features extracted by the model itself~\cite{byol, jepa}. This ability of the model to determine its own latent representation for making predictions can lead to representation collapse.
Authors in~\cite{combination} have found that the combination of latent predictions and hebbian learning is able to circumvent this issue.
They have demonstrated its performance mainly for non-spiking neural networks, with only limited artificial experiments for SNNs.

\begin{figure*}
\centering
\includegraphics[width=6in]{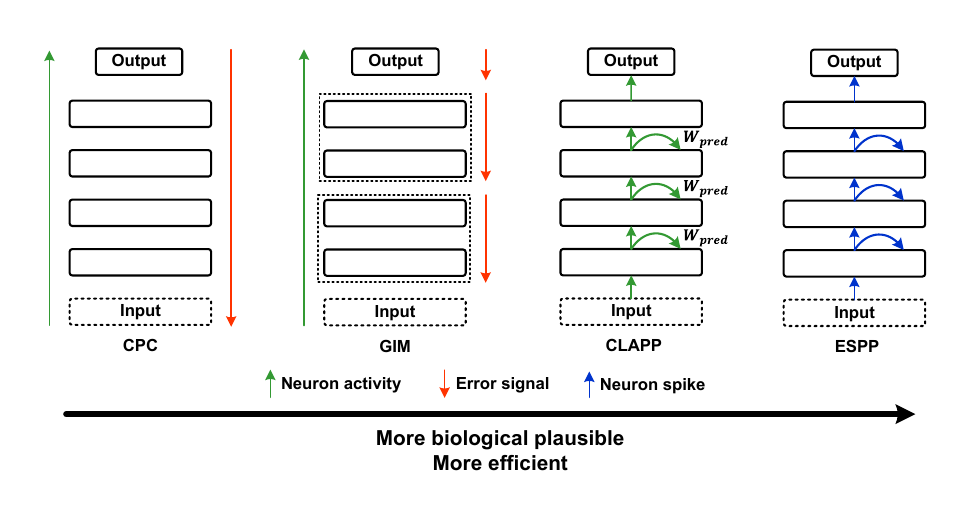}
\caption{EchoSpike Predictive Plasticity (ESPP) Compared to Related Work}
\label{fig:espp}
\end{figure*}

A second local learning rule inspired by SSL is CLAPP (Contrastive Local and Predictive Plasticity)~\cite{clapp}.
CLAPP has its origins in contrastive predictive coding (CPC)~\cite{cpc}.
The essence of CPC lies in its ability to learn rich representations by maximizing the mutual information between current and future latent space representations.
This is achieved through a contrastive learning framework, where the model is trained to differentiate between the actual future latent representation and a set of negative samples, which are representations drawn from different contexts.
Authors in~\cite{gim} have worked towards making CPC more biologically plausible by eliminating the need for end-to-end backpropagation.
This adaptation is achieved through the introduction of Greedy InfoMax (GIM), a method that localizes the learning process within each module of the network.
Building on this, CLAPP represents the next step in this evolution.
CLAPP integrates the principles of GIM and derives from it a local self-supervised learning rule for ANNs.
Figure~\ref{fig:espp} illustrates the evolution of learning algorithms from CPC to ESPP.

Although ESPP is inspired by CLAPP, there are notable differences between the two:
\textbf{(1)} CLAPP utilizes an additional recurrent weight matrix per layer for making predictions, whereas ESPP does not require these extra prediction matrices.
\textbf{(2)} CLAPP predicts future activity within a sample and adapts the forward weight matrices such that the activity either matches these predictions, or is far away from these predictions, depending on whether they changed the sample or not.
In contrast, ESPP uses the entire previous sample, termed as the ``echo'', as a prediction. It then adapts the forward weights to ensure the current activity is either close to the echo or far from it, based on whether the label corresponding to the echo was the same or different.
\textbf{(3)} ESPP is specifically developed for SNNs, whereas CLAPP only works for standard ANNs.

A third learning rule that resembles ESPP in some ways is the Forward-Forward algorithm~\cite{hinton2022forward}.
The Forward-Forward algorithm uses layer-wise objectives, and tries to achieve high goodness for positive samples and low goodness for negative samples.
ESPP also uses layer-wise objectives, but in contrast to the Forward-Forward algorithm, it compares pairs of samples to each other in a way that is more in line with CPC.
\subsection*{Contributions of this work}
This work provides five significant contributions to the state-of-the-art:

\textbf{1. Leveraging on Local Targets in ESPP:} Instead of relying on global targets and approximating backpropagation of the corresponding errors, we take inspiration from the field of SSL and optimize a local (layer-wise) target. This approach not only facilitates superior scalability and the development of hierarchical features, but also uniquely enables the effective use of all neural network layers in downstream classification tasks, rather than relying solely on the last layer.

\textbf{2. Empirical Validation of Performance and Scalability:} ESPP demonstrates competitive performance and scalability, validated through experiments on the N-MNIST~\cite{nmnist} and SHD~\cite{heidelberg} datasets.
It not only outperforms existing local learning rules for SNNs but also showcases remarkable adaptability across architectures, including multi-layer configurations up to four layers, recurrent connections, and skip connections.

\textbf{3. Innovative Approach to Event-based Weight Updates:} ESPP introduces a novel methodology to implement event-based weight updates, intelligently choosing the time steps that count the most.

\textbf{4. Intrinsic Regularization of Spiking Activity:} A notable feature of ESPP is its ability to intrinsically regulate spiking activity.
We analyze how this regularization works, and how we can use it to drive spiking activity to a desired level.

\textbf{5. Optimized for Neuromorphic Hardware Efficiency:} The ESPP algorithm is precisely engineered for neuromorphic hardware compatibility, promoting seamless on-chip learning without necessitating a synapse-scaling buffer.
This optimization ensures ESPP's practicality and broad applicability in real-world neuromorphic applications.

\section{Derivation of the ESPP learning rule}\label{sec:ESPP}
As ESPP is inspired by the CLAPP learning rule~\cite{clapp}, we first present the CLAPP Loss for layer l at time t as defined in~\cite{clapp}:
\begin{equation}
\begin{split}
\mathcal{L}^{t,l}_{CLAPP} &= {max}(0, 1 - y^t \cdot u^{t+\delta t, l}_t) \\
\mbox{with } y^t &=
\begin{cases}
    +1 & \text{for fixation} \\
    -1 & \text{for saccade}
\end{cases}
\end{split}
\label{eq:clapp}
\end{equation}
Where $ u^{t+\delta t, l}_t $ is the dot production of the activity of layer l ($\boldsymbol{z}^{t+\delta t, l}$)
and the prediction from the previous time step ($ \boldsymbol{\hat z}^{t,l} $).
In~\cite{clapp} the authors arrived at the prediction by linear projection of the activity of the same layer at the previous time step.
This required them to resort to two weight matrices per layer: one feed-forward matrix and one recurrent matrix for the predictions.
In contrast to CLAPP, ESPP did not use an additional weight matrix $W_{pred}$ to calculate the predictive feedback.
Rather, the accumulated spiking activity of the previous sample was used directly as a prediction.
This echoing of the activity of the previous sample is what inspired the name of EchoSpike.
Therefore: $ \boldsymbol{\hat z}^{t,l} = \boldsymbol{z}^{t, l}$.
This simplifies the learning rule and saves a significant amount of resources for neuromorphic implementation.

While CLAPP predicts individual time steps in a sample, ESPP tries to produce similar activity for consecutive samples with the same label and different activity for samples with different labels.
If the previous sample was different from the current sample (``saccade''), optimizing the loss leads to a representation that is distinguishable from the previous one.
If the previous sample was the same as the current sample (``fixation''), it produces similar representations.

This leads us to the following definition of the loss for ESPP:
\begin{equation}
\begin{split}
\mathcal{L}^{t,l}_{ESPP} &= {max}(0, \Tilde{c}(y) - y \cdot \boldsymbol{s}^{t,l} \cdot \boldsymbol{\Bar{s}}^l_{prev}) \\
\mbox{with } y &=
\begin{cases}
    +1 & \text{for fixation} \\
    -1 & \text{for saccade}
\end{cases}
\end{split}
\label{eq:loss}
\end{equation}
and
\begin{equation}
\Tilde{c}(y) = c(y) \cdot \Bar{i}^t
\end{equation}
Where $\boldsymbol{s}^{t,l}$ is the spiking activity of layer l at time step t and $ \boldsymbol{\Bar{s}}^l_{prev} $ is the normalized summed spiking activity of the previous sample.
The normalized summed spiking activity is calculated as follows:
\begin{equation}
\boldsymbol{\Bar{s}}^l_{prev} = \frac{\sum_t \boldsymbol{s}^{t,l}_{prev}}{n_{tot}}
\end{equation}
$n_{tot}$ is the total number of spikes of layer l over all time steps of the previous sample.
The constant ``1'' in the original CLAPP loss function (Equation~\ref{eq:clapp}) was replaced by an adaptive threshold $ \Tilde{c}(y) $ for two reasons.
First, in contrast to ANN activity, spikes are always one or zero, and therefore the value of the dot product $\boldsymbol{s}^{t,l} \cdot \boldsymbol{\Bar{s}}^l_{prev}$, which we call \textit{similarity score}, can not be scaled arbitrarily.
The hyperparameters $c(y)$ that determine the threshold scale are crucial for ESPP. For $y=1$, $c(y)$ is positive and for $y=-1$, $c(y)$ is negative.

Second, for temporal data, the amount of activity that is present at a certain time step can vary considerably depending on how much input there is.
This is the reason to make the threshold adaptive, and linear to the average number of input spikes to the network $\Bar{i}^t$.
This scalar value is the same for all layers.
It is not dependent on the layer-wise input, but the input to the entire SNN, in order to stabilize activity over multiple layers.

\begin{figure*}[!t]
\centering
\subfloat[Fixation Example]{\includegraphics[width=3.5in]{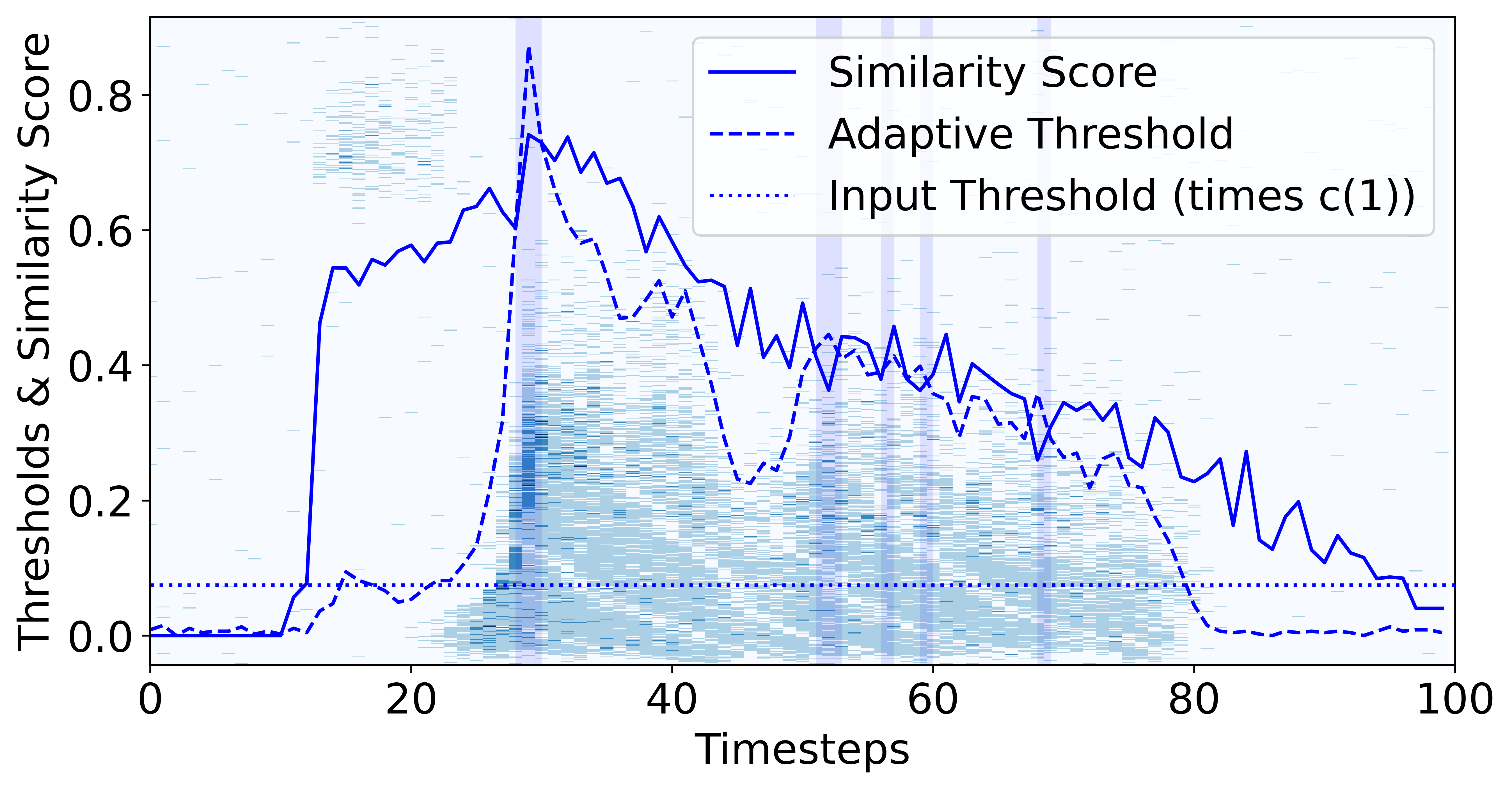}
\label{fig:adaptive_predictive}}
\hfil
\subfloat[Saccade Example]{\includegraphics[width=3.5in]{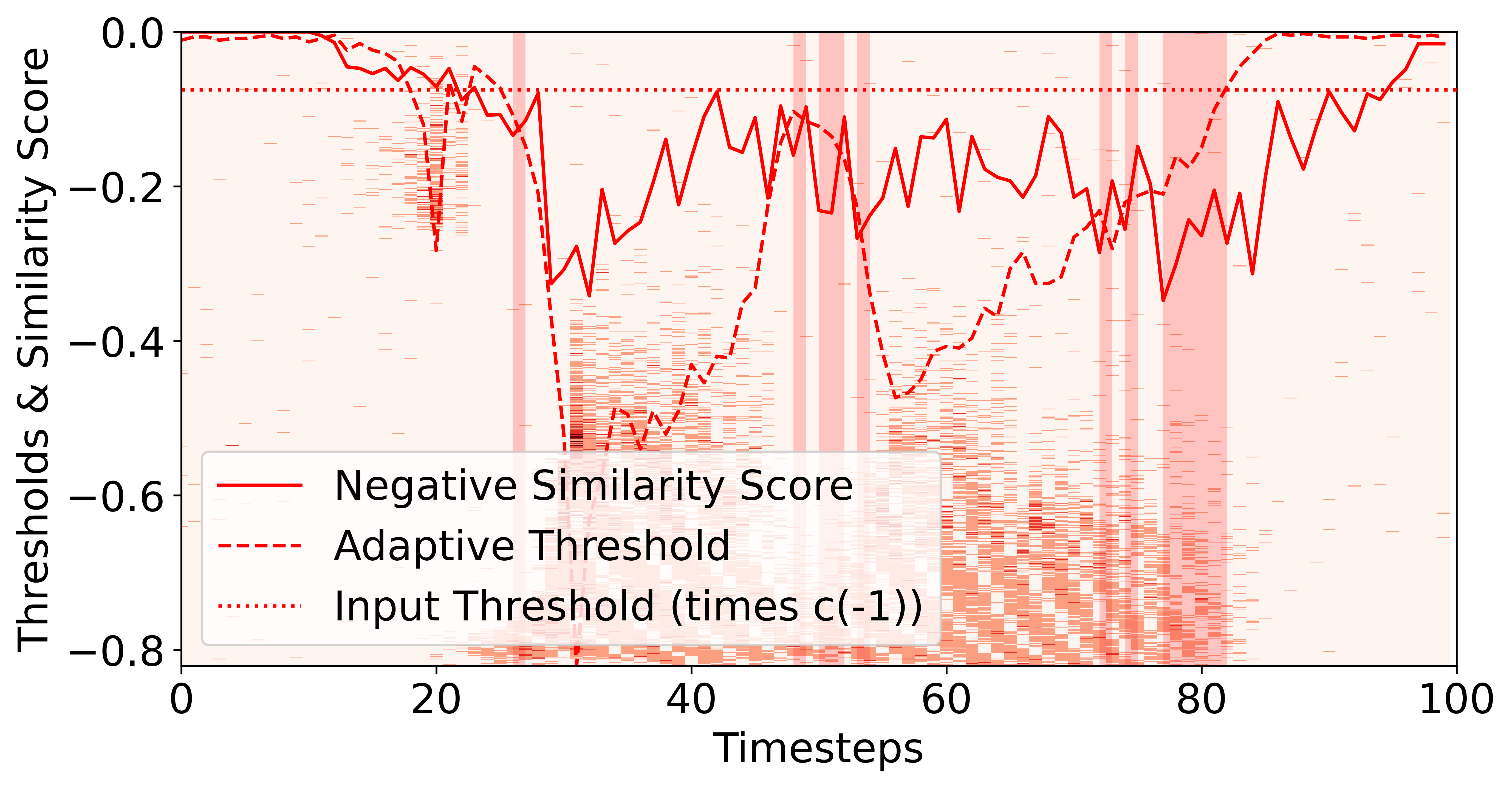}
\label{fig:adaptive_contrastive}}
\caption{a) An illustration of the adaptive threshold mechanism.
The background is filled with the audio data of a spoken digit (“Seven”) from the SHD dataset~\cite{heidelberg}.
The three lines indicate the value of the adaptive threshold $\Tilde{c}(y)$, the similarity score ($\boldsymbol{s}^{t,l} \cdot \boldsymbol{\Bar{s}}^l_{prev}$) and the input threshold $i_{thr}$.
The input threshold has been scaled by $c(y)$, so that it can be compared to the adaptive threshold.
If the adaptive threshold is below the input threshold, no weight update will be performed.
Time steps with a filled background indicate those time steps where a weight update will be performed.
b) Similar to a), but for a saccade.
Since the previous sample has been of a different label, the similarity score should be as small as possible.
For illustration purposes, the negative of the similarity score is plotted.
The adaptive threshold and the scaled input threshold are negative, because $c(-1)$ is negative.}
\label{fig:adaptive}
\end{figure*}

Now the derivative of the above loss function with respect to the feed-forward weights of layer l is following:
\begin{equation}
\frac{\partial \mathcal{L}^{t,l}_{ESPP}}{\partial W_l} = -y \cdot {dL}^t \cdot \frac{\partial (\boldsymbol{\Bar{s}}^l_{prev} \cdot \boldsymbol{s}^{t,l})}{\partial W_l}
\end{equation}
dL is the derivative of the outer function of the loss.
As a necessary approximation, we regard $\boldsymbol{\Bar{s}}^l_{prev}$ as a constant with respect to the weight matrix.
For the derivative of $\boldsymbol{s}^t$, we use a non-vanishing surrogate gradient function and eligibility traces~\cite{superspike, eprop}.
We use the derivative of the arc tangent function as our surrogate gradient.
The presynaptic eligibility trace $\boldsymbol{\tau}^{t,l}_i$, which quantifies the effect of each input to the current activity, is calculated as follows:
\begin{equation}
\boldsymbol{\tau}^{t,l}_i = \beta \boldsymbol{\tau}_i^{t-1, l} + \boldsymbol{s}^{t,l-1}
\end{equation}
Where $\beta$ is the decay constant.
The same decay was used for the eligibility trace as for the membrane potential of the leaky integrate and fire (LIF) neurons as introduced in previous SNN algorithms~\cite{eprop, superspike}.
With these terms, the derivative can be written as:
\begin{equation}
\frac{\partial \mathcal{L}^{t,l}_{ESPP}}{\partial W_l} \approx -y \cdot {dL}^{t,l} \cdot ({surr}(\boldsymbol{V}^t) \odot \boldsymbol{\Bar{s}}^l_{prev}) \otimes \boldsymbol{\tau}_i^{t, l}
\label{eq:grad}
\end{equation}
Here, $\otimes$ is the outer product and $\odot$ element-wise multiplication.
The derivative of the outer function can be calculated as follows:
\begin{equation}
{dL}^{t,l} = \begin{cases}
    1 & \text{if } \Tilde{c}(y) \geq y \cdot \boldsymbol{s}^{t,l} \cdot \boldsymbol{\Bar{s}}^l_{prev} \And \bar{i}^t \geq i_{thr}\\
    0 & \text{else}
\end{cases}
\label{eq:dL}
\end{equation}
This outer derivative can be thought of as a gating mechanism for the weight updates.
In addition to the adaptive activity threshold, a threshold of the input activity ($i_{thr}$) is added, which is not derived from the loss function. If this threshold is not reached, no weight update will be performed. The model only updates the weights to meaningful input, and not to background noise in this way.
Additionally, the input threshold leads to an increase in efficiency, especially when the input is temporally sparse.
Figure~\ref{fig:adaptive} illustrates the thresholds and the weight update gating mechanism, plotting the relevant values over time for a fixation and saccade sample.

\subsection{\textbf{Online Local Learning}}
Because ESPP is an online learning algorithm, weight updates are performed at every time step.
The weight update consists of the negative derivative of the loss with respect to the weights (Equation~\ref{eq:grad}) multiplied by a learning rate.
As can be seen from the different components of the gradient: None of the terms require a buffer that scales with the number of connections.
All the terms are local in time and space and are suitable for on-chip learning on neuromorphic hardware.

Compared to the other components, ${dL}^t$ is only local per layer, rather than local per neuron.
This could be omitted by adding another neuron per layer that performs the layer summation of ${dL}^t$.
In order for the spiking activity of this ${dL}$ neuron to be equivalent to ${dL}^t$, the weights of that neuron would have to be set to $-y \cdot \boldsymbol{\Bar{s}}^l_{prev} $ and the threshold to $\Tilde{c}(y)$.
This way, we can exploit the similarity between ${dL}^t$ and a spiking neuron.
The activity of this neuron has to be fed back to all the other neurons in the same layer and will trigger a weight update.
This feedback could be compared to biological neurons, where top-down feedback signals to the apical dendrites of cortical pyramidal neurons have been linked to solving the credit assignment problem in BNNs~\cite{Richards_Lillicrap19, Sacramento_etal18}.

\subsection{\textbf{Intrinsic Regularization}}
One challenge with training SNNs, especially with local learning, is to keep the spike rates in a useful range.
If $c(y)$ is not set properly, it can happen that the network converges to no spikes at all, at which point all weight updates will be zero, and the network has essentially died.
Or, it can happen that all neurons fire at every time step.
With properly set $c(y)$, ESPP has an intrinsic way to regularize the number of spikes to an ideal spike rate.

The surrogate gradient, spikes, and spike traces are always larger or equal to zero.
In Equation~\ref{eq:grad}, we can see that, therefore, for fixations all weight updates will always be larger than or equal to zero, since the gradient is always negative or zero.
For saccades, the opposite is the case: weight updates will always decrease the weights, leading to less spiking activity.
Therefore, we have the dynamics that predictive updates (fixation) increase spiking activity and contrastive updates (saccade) decreases spiking activity.

${dL}$ is either zero or one, depending on whether the ESPP loss was sufficiently small.
A high number of spikes makes it hard for the contrastive loss to be below the threshold ($c(-1)$), because each spike can only make the loss larger for a saccade (Equation~\ref{eq:loss}).
For a fixation, exactly the opposite is the case.
More spikes make it easier to have a loss below the threshold.
This leads to ${dL}$ being zero for more samples (Equation~\ref{eq:dL}), which in turn leads to fewer predictive updates being added to the weight matrix (compared to contrastive updates), which leads to less activity.
This forms a negative feedback loop that pulls the spiking activity back to its optimum, if it gets too large.
Too few spikes lead to more predictive updates, as contrastive updates, which will increase the spiking activity, until it is back at a balance point.

These two feedback loops lead to a stable, optimal sparsity.
By setting $c(y)$ appropriately, we can influence, how sparse the solution should be and how clearly separable spiking activity of different labels should be.
\subsection{\textbf{Sparse Weight Updates}}\label{sec:sparsity}
Previous event-based learning rules, such as ETLP, do not adaptively select at which time steps a weight update should be performed.
They either use regular intervals, or an additional external algorithm, to determine which time steps lead to an update.

For ESPP, the thresholds ($\Tilde{c}(y)$ and $i_{thr}$) have the effect that the weight updates are not performed at every time step.
By setting these thresholds accordingly, we can influence the percentage of weight updates that are executed.
In contrast to standard event based updates, ESPP intrinsically has the ability to selectively choose those time steps that matter the most.

For the results reported in our best performing model, approximately 18\%-27\% (decreasing with training) of weight updates were performed.
An improvement to the current algorithm could attempt to set $c(y)$ adaptively during training.
Given a target weight update sparsity and a target average firing rate, this adaptivity would set $c(y)$ such that these targets are met during training.
This would reduce the effort of finding suitable hyperparameters, whilst simultaneously providing better control over the sparsity of weight updates and the firing rate of neurons.
\subsection{\textbf{Generalization to Different Architectures}}
The algorithm described above can be used without any altercation for RSNNs of variable depth. We can simply concatenate the activity of the previous time step to the input of the same layer (or any of the previous layers). This way, the input vector and presynaptic trace get larger, but everything else stays the same from the perspective of an EchoSpike layer. A recurrent connection from the last layer back to the first layer (deep transition RNN~\cite{pascanu2013construct}) can also be added, instead of adding recurrence to every layer. Similarly, skip connections by concatenating the activity of a layer (or the input to the network) to the input of a deeper layer are also possible for ESPP. It remains an open research question to what types of additional architectures ESPP can successfully be applied.
\section{Low Cost Supervised Learning for Classify Layer}\label{sec:supervised}
EchoSpike Predictive Plasticity on its own does not predict anything.
To make predictions, we use a two phase local learning approach.
In a first phase, ESPP is used to train the hidden layers by finding features that differ as much as possible for different classes, and as little as possible for same classes.
In a second phase, we capitalize on these features by training a low-cost classifier that connects these features to the correct label, without changing the hidden layers.

We propose three different types of classifiers.
A classifier trained with gradient descent, a closed-form classifier and classification via few-shot learning.
\subsection{\textbf{Gradient Descent Classifier}}\label{sec:GD}
The output layer of our gradient descent classifier consists of $n_{classes}$ non-leaky (decay $=1$), non-spiking integrator neurons.
The weights are updated via a simple learning rule at the last time-step of each sample.
The softmax of the membrane potential at the final time step is seen as the prediction of the network.
By using the derivative of the cross-entropy loss, combined with the non-decaying input trace $\boldsymbol{\tau}_i$ (which is just the sum of the input vectors), we get the following weight update rule:
\begin{equation}
{dW}_{out} = (\boldsymbol{l}_{onehot} - \boldsymbol{p}) \otimes \boldsymbol{\tau}_i
\end{equation}
Where the prediction $\boldsymbol{p}$ is the softmax of the membrane potential at the last time step ($\boldsymbol{V}^{t_{last}}$) and $\boldsymbol{l}_{onehot}$ is the one-hot encoded label.
In contrast to the hidden layers, where stochastic gradient descent was used, we used Adam~\cite{kingma2014adam} to optimize the classifier.

Because there is no decay in the output layer, multiplying the inputs at each time step with the weight matrix is not needed.
First accumulating the inputs and multiplying it with the weight matrix only at the last time step is also an option.

\subsection{\textbf{Closed-Form Classifier}}\label{sec:closed}
The gradient descent classifier explained above can be reduced to a simple linear regression with the following two small changes: (1) exchange the cross-entropy loss with an L2 loss; (2) remove the softmax activation.
The classifier simply has to predict the one-hot encoded labels by multiplying the accumulated inputs of the last time step of each sample with a single weight matrix.
This linear regression can be solved in closed-form, without the need to train an output layer with gradient descent.
This approach was tested on the SHD dataset~\cite{heidelberg} (Section~\ref{sec:shd}) and the results can be found in Figure~\ref{fig:shd_ff} and~\ref{fig:shd_stacked}.

\subsection{\textbf{Few-Shot Learning}}\label{sec:fewshot}
A third method we explored was predicting the class label with few-shot learning without gradient decent.
For this approach, twenty training samples were randomly selected from each class and fed into pretrained hidden layers.
The spiking activity of each hidden layer was recorded, summed, and normalized to obtain one reference vector per layer and per class.

These vectors can then be used to generate predictions in the following way:
For each sample in the test set, sum the activity over time for each layer and compare it with the reference vectors of each class.
The class whose reference sample is closest to the activity of the current sample is chosen as the prediction.
Since the hidden layers are trained to minimize the ESPP loss, we define the closeness as having the lowest ESPP loss (for $y=1$), where the reference vector of each class is used as the echo.

In this way, the test accuracy can be obtained for each layer separately.
In practice, one would do it for the last layer only or for the activity of all layers combined.
This approach is extremely low cost and worked well for N-MNIST~\cite{nmnist}.
For SHD, which has more temporal information, competitive results could not be achieved with this method.
The results for few-shot learning on N-MNIST can be found in Figure~\ref{fig:NMNIST_online_fewshot}.
\subsection{\textbf{Connecting Hidden Layers to the Output Layer}}\label{sec:connect}
In multi-layer neural networks trained with backpropagation, all neurons in every layer are optimized on the same global goal.
Since neurons in early layers only serve the purpose of helping later layers in solving this global goal, it is most likely not useful to connect early hidden layers directly to the output layer.
Instead, usually only the last layer is connected to the output layer.

In contrast, ESPP uses local (layer-wise) targets.
In a feed-forward neural network, each layer tries to find features that are present in samples of the same class but different in samples of different classes.
Each layer builds these features with the features that it received from the previous layer.
This is how a multi-layer SNN trained with ESPP builds hierarchical features.

In the supervised learning phase of the two-phase approach described before, we explore two options to connect the hidden neurons to the output layer.
The first approach is the standard way: connect only the last hidden layer to the output layer.
We call this approach \textit{last-layer-only prediction}.
The second approach we used is to connect all available layers (all the hidden layers and the input layer) to the output layer.
We call this approach \textit{all-layers prediction}.
For SHD the all-layers prediction clearly improved the performance, whilst for N-MNIST there was no clear benefit.
This is why the \textit{last-layer-only prediction} is used for N-MNIST, and a comparison between the two approaches for SHD can be found in Section~\ref{sec:shd}.
\section{Empirical results}\label{sec:results}
We employed the ESPP algorithm on two common neuromorphic benchmarks to demonstrate its performance compared to other algorithms such as ETLP, backpropagation through time (BPTT) and eProp.
We implemented the algorithm explicitly with PyTorch and snnTorch~\cite{snntorch}, without the use of automatic differentiation.
The code is publicly available on GitHub \footnote{\url{https://github.com/largraf/EchoSpike}}.
The most important hyperparameters are listed in Table~\ref{table:hyper}.
Even tough, the online learning weight update rule is performed at every time step, we still used a batch size of 128, meaning that the weight update rule was performed at every time step, simultaneously for 128 samples that are run in parallel.
This was necessary to speed up computation on the GPU.
\begin{table}[h!]
\centering
\caption{Hyperparameters for N-MNIST and SHD}
\begin{tabular}{||c | c | c||}
 \hline
 Parameter & N-MNIST~\cite{nmnist} & SHD~\cite{heidelberg} \\ [0.5ex]
 \hline\hline
 Number of Time Steps & 10 & 100 \\
 Hidden Neurons per Layer & 200 & 450 \\
 Learning Rate & 1e-4 & 1e-4 \\
 Epochs  & 300 & 1000 \\
 Batch Size (Samples in Parallel on GPU) & 128 & 128 \\
 Batch Size in Time & 1 & 1 \\
 Input Threshold ($i_{thr}$) & 2\% & 5\% \\
 Predictive Threshold Constant ($c(1)$) & 2 & 1.5 \\
 Contrastive Threshold Constant ($c(-1)$) & -1 & -1.5 \\
 Decay Constant ($\beta$) & 0.9 & 0.95 \\[1ex]
 \hline
\end{tabular}
\label{table:hyper}
\end{table}
\subsection{\textbf{N-MNIST}}\label{sec:N-MNIST}
Since N-MNIST~\cite{nmnist} lacks extensive temporal data, a 3-layer feed-forward architecture was utilized for this dataset. Each layer consists of 200 LIF neurons.

\subsubsection{\textbf{Gradient Descent}}
To demonstrate the classifier's minimal sample requirement, the training data was split into a 90/10 ratio. The 90\% portion was used for training the hidden layers with ESPP, while the remaining 10\% was used for training the classifier with supervised learning. The hidden layers were trained for 300 epochs and the classifier for just 1 epoch.
The output layer training was repeated 10 times using the gradient descent method described in Section~\ref{sec:GD}, both from each layer (\textit{last-layer-only prediction}) and directly from the inputs. The mean train and test accuracies, along with the standard deviation, are presented in Figure~\ref{fig:NMNIST_online_acc}.

We can see for N-MNIST that the performance of the gradient descent classifier does not increase significantly with more than one hidden layer.
The performance of the classifier from all three hidden layer was almost identical (mean test accuracy of Layer 1, 2, 3: 95.28\%, 95.43\%, 95.35\%).
For comparison to ETLP, BPTT and eProp, the results for 1 hidden layer was used because it uses the same number of trainable parameters.

As demonstrated in Table~\ref{table:N-MNIST}: ESPP compares favorably to ETLP, even tough we only used 10\% of the training data to train the classifier and 90\% of the training data for training the hidden layers with ESPP.
The results reported by~\cite{ETLP} also use 200 hidden LIF neurons and therefore the same number of trainable parameters as our result from layer 1.
This clearly demonstrates that a competitive classifier can be trained with very low cost, after the hidden layers have been trained with ESPP.
\subsubsection{\textbf{Few-shot learning}}
For the few-shot learning experiment, we proceeded as explained in Section~\ref{sec:fewshot}.
The experiment was repeated 10 times, in order to measure how sensitive this is to the exact samples that have been selected from the 10\% of the training set.
We found that using fewer than 20 samples per class leads to a large variance, because it depends more on which samples are chosen.

\begin{figure*}
\centering
\subfloat[Gradient Descent Classifier]{\includegraphics{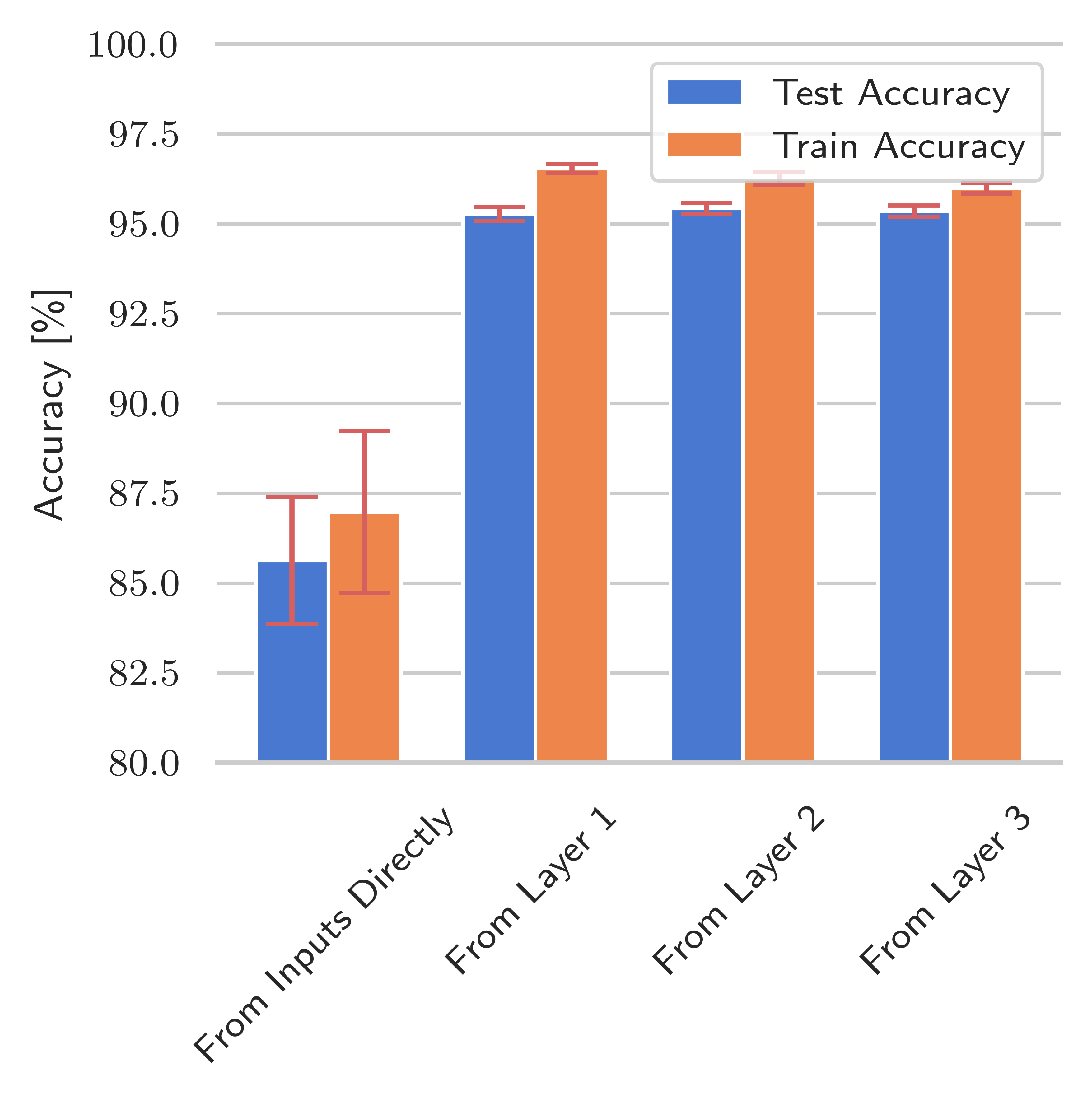}
\label{fig:NMNIST_online_acc}}
\hfil
\subfloat[Few-Shot Learning]{\includegraphics{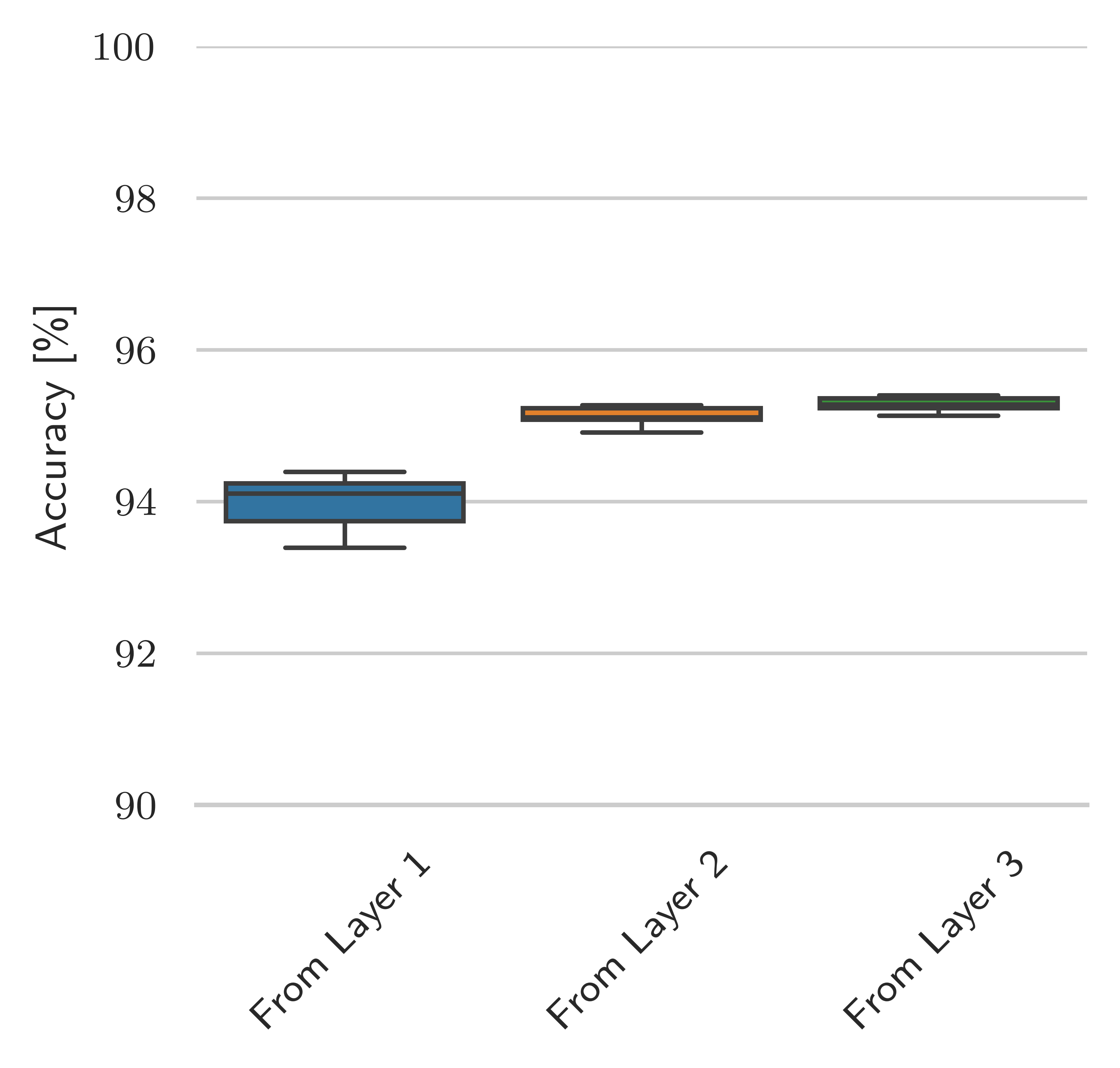}
\label{fig:NMNIST_online_fewshot}}
\caption{N-MNIST train and test accuracy for the gradient descent classifier (a) and from the few-shot learning experiment (b), trained for each layer separately.}
\end{figure*}

The results can be seen in Figure~\ref{fig:NMNIST_online_fewshot}.
For the few-shot experiments, there seems to be improved performance with depth.
A possible reason is that deeper layers have more clearly separable representations, which is not necessary for a classifier trained with gradient descent, but more important for few-shot learning.
The few-shot accuracy from layer 3 is the best and has an average accuracy of 95.68\%.
This result is above the ETLP result~\cite{ETLP}, which had an accuracy of 94.30\% for a network with only one hidden layer, and also slightly better
than the test accuracy of our best classifier trained with gradient descent.
\begin{table*}
\centering
\begin{threeparttable}
\caption{Comparison to SoTA results on N-MNIST~\cite{nmnist}}
\label{table:N-MNIST}
\begin{tabular}{||c c c c c c||}
 \hline
 Model & Local\tnote{2} & Classifier Type & \# Hidden Layers & \# Parameters & Test Accuracy (\%) \\ [0.5ex]
 \hline\hline \rule{0pt}{2.5ex}
 ETLP~\cite{ETLP} & \cmark & & 1 & 416.8k & 94.30 \\
 BPTT~\cite{ETLP} & \xmark & & 1 & 416.8k & 97.67 \\
 eProp~\cite{ETLP} & \xmark & & 1 & 416.8k & 97.90 \\
 \hline \rule{0pt}{2.5ex}
 ESPP (ours) & \cmark & Gradient Descent\tnote{1} & 1 & 416.8k & 95.28 \\
 ESPP (ours) & \cmark & Few-Shot w/o Gradient Descent  & 1  & 416.8k &  93.90 \\
 ESPP (ours) & \cmark & Few-Shot w/o Gradient Descent & 2  & 456.8k &  95.15 \\
 ESPP (ours) & \cmark & Few-Shot w/o Gradient Descent & 3  & 496.8k & 95.68 \\ [0.5ex]
 \hline
\end{tabular}
\begin{tablenotes}
\item [1] using only 10\% of the training data for 1 epoch
\item [2] local in time and space
\end{tablenotes}
\end{threeparttable}
\end{table*}
\subsection{\textbf{SHD}}\label{sec:shd}
The SHD dataset~\cite{heidelberg} consists of auditory data of spoken digits from zero to nine, both in German and in English (20 different classes).
Since it consists only of 8156 training samples, the whole training dataset was used both for training the hidden layers with ESPP and for training the classifier.
We trained a 3 layer feed-forward SNN and a 3 layer stacked RSNN, and
tried both: the \textit{last-layer-only prediction} and the \textit{all-layers prediction} approach described in Section~\ref{sec:connect}.
We also compared two different types of classifiers: the gradient descent classifier and the closed-form classifier, as described in Section~\ref{sec:supervised}.
Overfitting is a large issue for SHD.
To counter overfitting, an additional 4 layer feed-forward SNN was trained with data augmentation, which resulted in our best performance.

\subsubsection{\textbf{Feed-Forward Network}}
We first tested the ESPP algorithm on the SHD dataset with a 3 layer feed-forward network.
Each layer contains 450 leaky LIF neurons.

\begin{figure*}
\subfloat[Feedforward]{\includegraphics{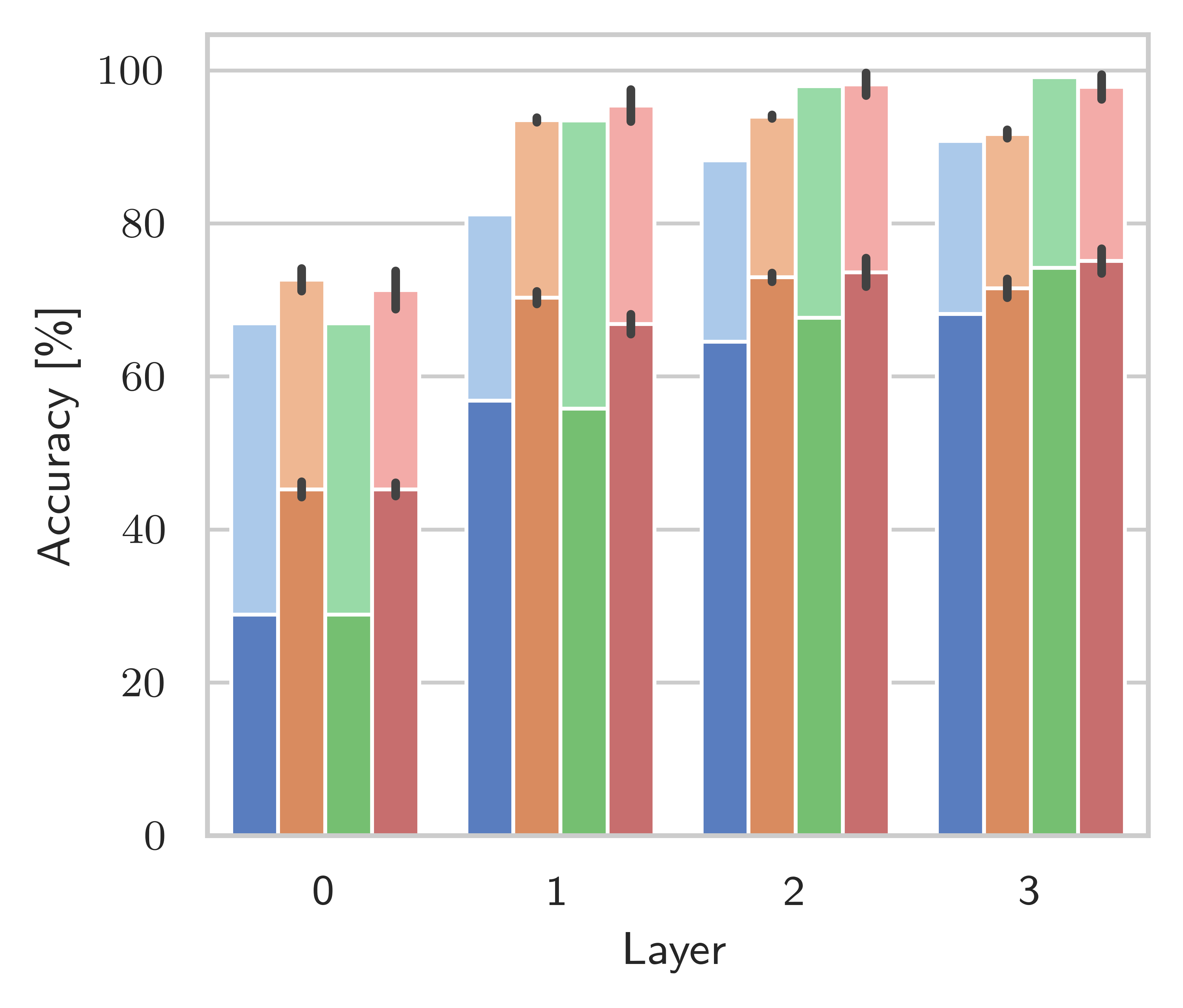}
\label{fig:shd_ff}}
\subfloat[Recurrent]{\includegraphics{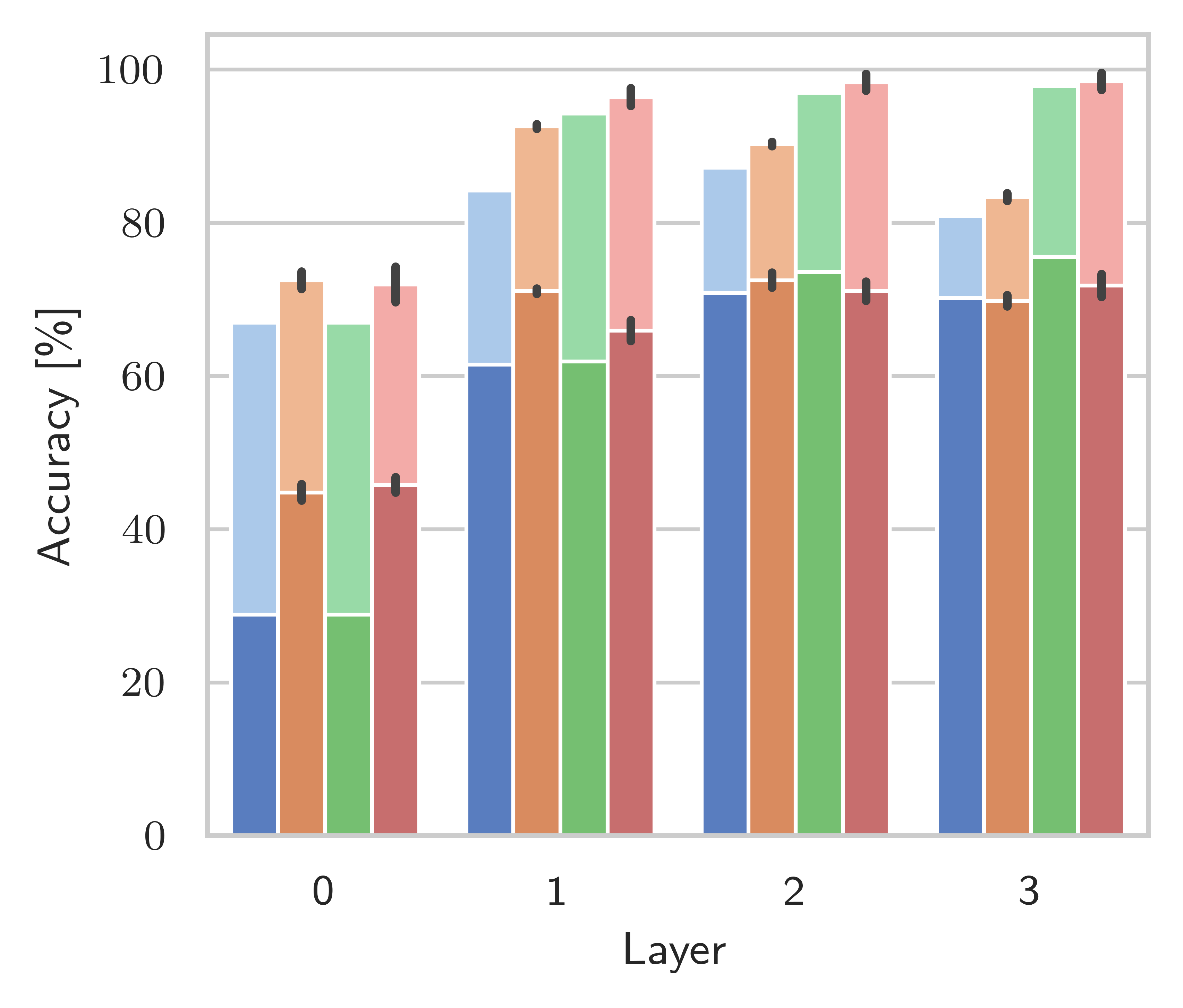}
\label{fig:shd_stacked}}

\centering
\subfloat{\includegraphics{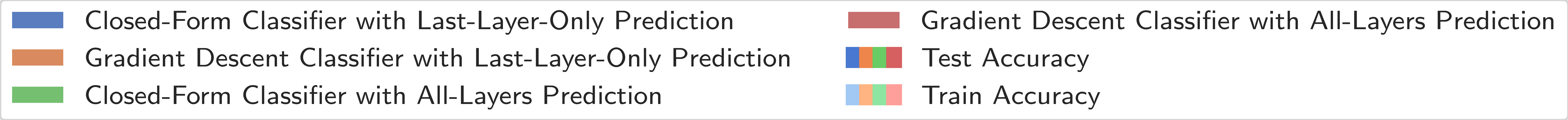}}
\caption{Train and test accuracies for the feedforward (a) and recurrent (b) architectures on the SHD dataset.
We compare the four possible combinations between the two different types of classifiers (gradient descent and closed-form) and the two different types of connecting the hidden layers with the output layer (last-layer-only prediction and all-layers prediction).}
\end{figure*}

The hidden layers of the network were trained with the ESPP algorithm for 1000 epochs.
As explained in Section~\ref{sec:connect}, we can connect the hidden layers to the output layer in two different ways: \textit{all-layers prediction} and \textit{last-layer-only prediction}.
Furthermore, two different types of classifiers were used (closed-form and gradient descent).
The train and test accuracies of all four possible combinations between the mentioned options are summarized in Figure~\ref{fig:shd_ff}.
Compared to N-MNIST, SHD was more challenging to classify with the gradient descent classifier, which is why 30 epochs were used, instead of only 1 epoch.
The gradient descent classifier was trained 10 times and the mean train and test accuracy is reported, as well as the standard deviation.

For all options, there is a large difference between train and test accuracy, which indicates strong overfitting on the training set.
The \textit{all-layers prediction}, in all cases, has a higher accuracy on the training set.
On the test set, the difference is not as clear anymore.
This is likely due to increased overfitting.
For one hidden layer, the \textit{last-layer-only prediction} performs better than the \textit{all-layers prediction}, which uses the inputs as well as the hidden layer spiking activity to predict the output.
This indicates that using the inputs directly increased overfitting, leading to a worse performance on the test set.
Overall, we can see that the \textit{all-layers prediction} scales better with depth.
Between the gradient descent and the closed-form classifier, the gradient-descent classifier mostly performs better.

The classification accuracy on the test set of 70.31\% from layer 1 (\textit{last-layer-only prediction}, gradient descent) clearly outperforms the accuracy of a 1 layer feed-forward network containing the same number of LIF neurons trained with ETLP~\cite{ETLP}, which only achieved a test accuracy of 59.19\% as shown in Table~\ref{table:shd}.
More surprisingly, it is also higher than the results from eProp (63.04\%) and BPTT (66.33\%) reported in the same paper~\cite{ETLP}.

To calculate the closed-form solution, we used the lstsq function from SciPy
\footnote{\url{https://docs.scipy.org/doc/scipy/reference/generated/scipy.linalg.lstsq.html}}, which computed in no more than a few seconds per classifier on a laptop CPU.
This again demonstrates that with ESPP competitive results can be achieved with low supervised learning cost.

\subsubsection{\textbf{Recurrent Network}}
In the RSNN each layer feeds its own spiking output back to itself with an additional weight matrix.
We stack three of these recurrent layers (each consisting of 450 LIF neurons) in series and perform the same experiment as with the feed-forward architecture.
The same four plots are shown in Figure~\ref{fig:shd_stacked}.
Compared to the feed-forward network, the RSNN has more trainable parameters for the same number of neurons.
Even tough the RSNN has more parameters, the performance is similar.
\subsubsection{\textbf{Adding Data Augmentation}} \label{sec:besteffort}
\begin{figure}
    \centering
    \includegraphics{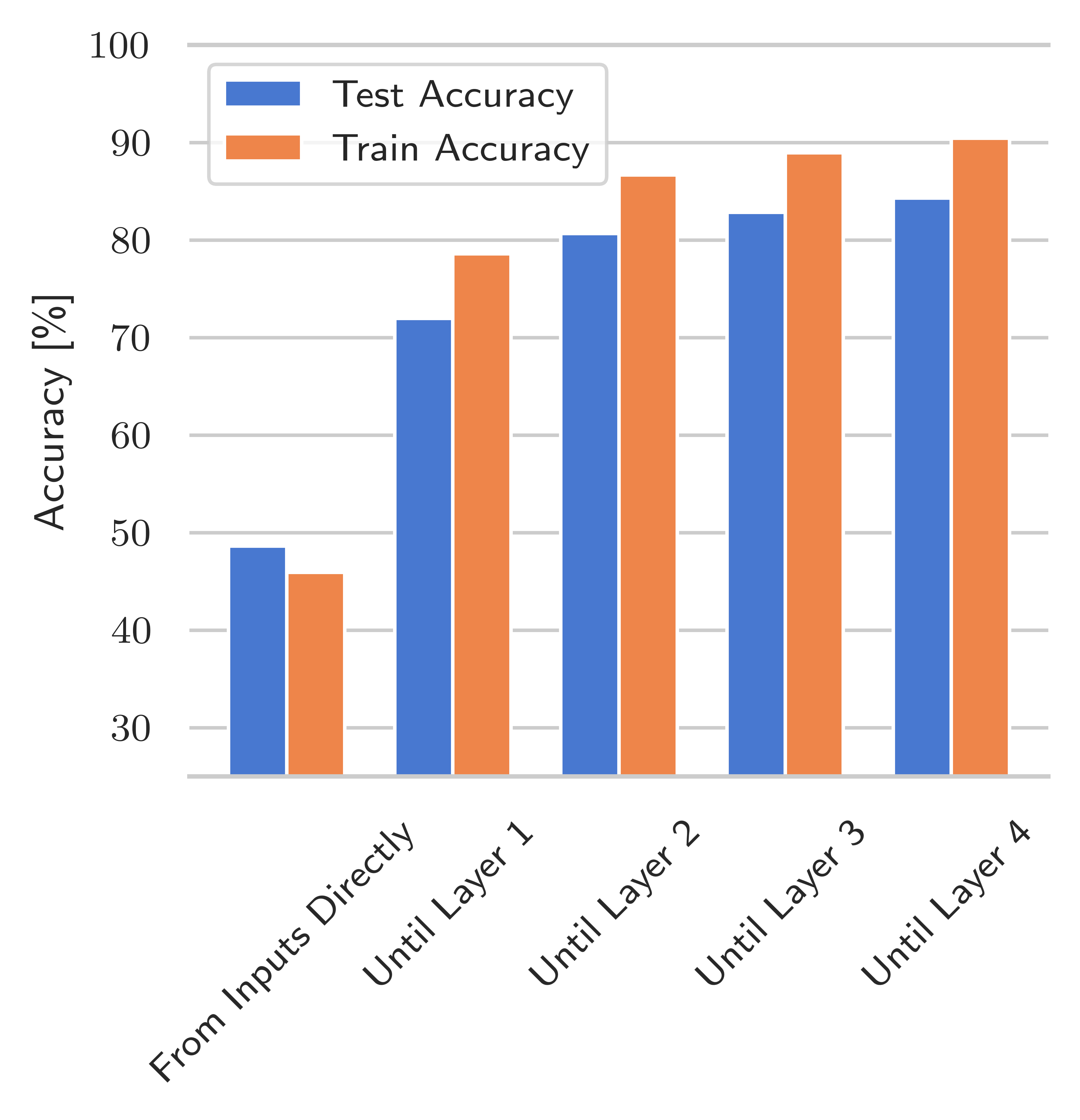}
    \caption{Train and test accuracy of a feed-forward neural network on SHD with data augmentation and gradient descent all-layers prediction.}
    \label{fig:best_effort}
\end{figure}

Due to the strong overfitting, an additional 4 layer feed-forward neural network with 450 LIF neurons per layer was trained.
\cite{nowotny2022loss} reported that, out of all types of data augmentation they tried, randomly shifting the sound data in the frequency direction had the biggest success.
Inspired by these results, we randomly shifted all of our training data by up to $\pm 35$ indices ($= \pm 5\%$) in the frequency direction.
The hidden layers were trained with ESPP for 1000 epochs.
The classifier was trained with gradient descent, the \textit{all-layers prediction} and the same type of data augmentation for 100 epochs.
More epochs were used for the supervised learning part than before, because with added data augmentation, it takes longer for the classifier to converge.
The performance can be seen in Figure~\ref{fig:best_effort}.
As expected, the data augmentation leads to lower performance on the training set and increased performance on the test set.
This model scales very clearly with depth.

As illustrated in Table~\ref{table:shd}, the best performance of a model trained with ETLP~\cite{ETLP}, is from an RSNN  with 450 LIF neurons with adaptive thresholds (ALIF), which had a test accuracy of 74.59\%.
The adaptive thresholds add additional complexity to the model, which makes it challenging for on-chip learning.
\cite{osttp} report a performance of 77.33\% for OSTTP.
OSTTP's  eligibility trace is the same as the one from OSTL~\cite{ostl}, which, due to its high dimensionality, 
is difficult to implement in its full form on a neuromorphic chip.
Even tough both, ETLP and OSTTP, add more complexity (ALIF neurons and complex eligibility traces) our model exceeds their performance with two hidden feed-forward layers. Our highest test accuracy of 84.32\% clearly outperforms all other local learning results (that we are aware of) on SHD.

It is also competitive with networks trained with BPTT.
As a comparison: \cite{heidelberg} reported the best effort result, for an RSNN with 1024 hidden neurons, of 83.2\% ($\pm 1.3\%$).
They also used data augmentation, had 84\% more trainable parameters than our best model, and used BPTT.
This clearly demonstrates that ESPP can compete with BPTT by using deeper networks.
\cite{heidelberg} reported that increasing the number of hidden layers for a feed-forward SNN did not lead to significant increase in test performance.
Our results on the other hand suggest that, by using ESPP and with the all-layers prediction, performance improves up to a depth of at least 4 hidden layers.
An overview of results on SHD can be found in Table~\ref{table:shd}.

\begin{table*}
\centering
\begin{threeparttable}
\caption{Comparison to SoTA results on SHD~\cite{heidelberg}}
\label{table:shd}
\begin{tabular}{||c c c c c c c c||}
 \hline
 Learning Rule & Local & Architecture & Classifier Type & \# Hidden Layers & \# Parameters & Neuron Type & Test Accuracy (\%) \\ [0.5ex]
 \hline\hline \rule{0pt}{2.5ex}
 ETLP~\cite{ETLP} & \cmark & Feed-Forward && 1 & 324k & LIF & 59.19 \\
 BPTT~\cite{ETLP} & \xmark & Feed-Forward && 1 & 324k & LIF & 66.33 \\
 ESPP (ours) & \cmark & Feed-Forward & Gradient Descent & 1 & 324k & LIF & 70.31 \\
\hline \rule{0pt}{2.5ex}
 ETLP~\cite{ETLP} & \cmark & Recurrent && 1  & 526.5k & ALIF\tnote{3} & 74.59 \\
 OSTTP~\cite{osttp} & \cmark &  Recurrent && 1 & 526.5k & SNU\tnote{3} & 77.33 \\
 ESPP\tnote{1} (ours) & \cmark & Feed-Forward & Gradient Descent\tnote{2} & 2 & 549.5k & LIF & 80.70 \\
 \hline \rule{0pt}{2.5ex}
  BPTT\tnote{1}~\cite{heidelberg} & \xmark & Recurrent && 1 & 1.72M & LIF & 83.2 \\
 ESPP\tnote{1} (ours) & \cmark & Feed-Forward & Gradient Descent\tnote{2} & 3 & 761k & LIF & 82.86 \\
 ESPP\tnote{1} (ours) & \cmark & Feed-Forward & Gradient Descent\tnote{2} & 4 & 972.5k &LIF & 84.32 \\ [0.5ex]
 \hline
\end{tabular}
\begin{tablenotes}
\item [1] with data augmentation
\item [2] using the all-layers prediction method
\item [3] larger computational or memory complexity
\end{tablenotes}
\end{threeparttable}
\end{table*}

\subsubsection{{\textbf{Deep vs Shallow Networks}}}
To further analyze the difference between multi-layer networks and networks with only one hidden layer, we trained a one wide hidden layer feed-forward neural network with the same number of parameters as our best performing 4 layer network.
This network uses 1332 hidden LIF neurons.
The network was trained the same way that the 4 layer feed-forward network with data augmentation was trained.
The test performance of this network was only 68.86\% and the training performance 78.47\%.
This performance is clearly worse than our best performing model (84.32\% test accuracy).
Increasing the size of the hidden layer did not increase the performance compared to a feed-forward network with one hidden layer of 450 LIF neurons.
\section{Discussion}
In this paper, we introduced ESPP, a novel approach to local learning in SNNs that leverages techniques from SSL.
Our method, inspired by CLAPP~\cite{clapp}, is adapted to the unique temporal dynamics of SNNs and enables efficient implementation on neuromorphic hardware.
We demonstrated how ESPP provides significant advancements over existing methodologies in terms of efficiency, scalability, and performance.

ESPP has shown superior performance on benchmark datasets such as SHD~\cite{heidelberg} and N-MNIST~\cite{nmnist} compared to existing local learning methods like ETLP~\cite{ETLP} and OSTTP~\cite{osttp}.
The improvements are not only in accuracy, but also in computational efficiency, making ESPP a viable option for real-world applications where resources are constrained.
By using a simple learning rule with low resource requirements, ESPP makes a compelling case for its suitability in neuromorphic hardware implementation.
This efficiency is crucial for deploying SNNs in edge devices, where power and computational resources are limited.

ESPP contributes to addressing one of the critical challenges in the broader adoption of SNNs — scalability.
By facilitating efficient and effective learning in deep SNN architectures, ESPP opens up new possibilities for complex, real-world applications that were previously out of reach.
The advancements presented in this paper have significant implications for the field of neuromorphic computing.
ESPP's efficiency and performance make it an attractive option for a range of applications, from edge computing to more complex cognitive tasks.

A notable advantage of ESPP is its inherent compatibility with unlabeled data, a feature that aligns well with the SSL paradigm.
ESPP's design allows it to leverage the rich, often untapped, reservoir of unlabeled data, making it particularly suitable for scenarios where labeled data is scarce or expensive to obtain.

In the results presented in this paper, ESPP's implementation relies on partial label knowledge to inform the neural network whether the current and previous samples are the same or different. Future work could explore the use of fully unlabeled data, potentially employing data augmentation and other techniques commonly used in the field of SSL.
Another promising direction is to develop stronger methods for selecting hyperparameters or to employ adaptive techniques that can fine-tune these parameters on the fly, as discussed in Section~\ref{sec:sparsity}.

Additionally, we could employ techniques, such as adaptive thresholds~\cite{eprop}, heterogeneous time constants~\cite{Perez-Nieves_etal21} or temporal delays~\cite{sun2023learnable, hammouamri2023learning}.
This has the potential to increase the networks performance at the cost of additional computational and memory complexity.
If we add heterogeneous time constants or synaptic delays to our neuron model, we could no longer use the same eligibility traces for all neurons of a specific layer, because now the influence that a presynaptic neuron has on each postsynaptic neuron is not homogeneous anymore.
This is why, after adding heterogeneous time constants or synaptic delays, we would require eligibility traces that scale with the number of synapses, as opposed to with the number of input neurons.
\section{Conclusion}
ESPP represents a significant step forward in the development of efficient, scalable, and effective local learning rules for SNNs.
Its ability to improve performance over multiple layers, while reducing computational requirements, makes it a promising approach for facilitating the practical application of SNNs in a wide range of domains.
As we continue to explore and refine this approach, we anticipate that ESPP will play a crucial role in the evolution of neuromorphic computing and the realization of its full potential.
%
%


\printbibliography

\newpage

\vspace{11pt}



\vfill

\end{document}